# IMPROVING A SEQUENCE-TO-SEQUENCE NLP MODEL USING A REINFORCEMENT LEARNING POLICY ALGORITHM

Jabri Ismail[1], Aboulbichr Ahmed[1] and El ouaazizi Aziza[1,2]

[1]Laboratory of Engineering Sciences (LSI),
Sidi Mohamed Ben Abdallah University, Fez, Morocco
[2]Laboratory of Artificial Intelligence, Data Sciences and Emergent Systems
(LIASSE), Sidi Mohamed Ben Abdallah University, Fez, Morocco

## ABSTRACT

*Nowadays, the current neural network models of dialogue generation(chatbots) show great promise for generating answers for chatty agents. But they are short-sighted in that they predict utterances one at a time while disregarding their impact on future outcomes. Modelling a dialogue's future direction is critical for generating coherent, interesting dialogues, a need that has led traditional NLP dialogue models that rely on reinforcement learning.*

*In this article, we explain how to combine these objectives by using deep reinforcement learning to predict future rewards in chatbot dialogue. The model simulates conversations between two virtual agents, with policy gradient methods used to reward sequences that exhibit three useful conversational characteristics: the flow of informality, coherence, and simplicity of response (related to forward-looking function). We assess our model based on its diversity, length, and complexity with regard to humans. In dialogue simulation, evaluations demonstrated that the proposed model generates more interactive responses and encourages a more sustained successful conversation.*

*This work commemorates a preliminary step toward developing a neural conversational model based on the long-term success of dialogues.*

## 1. INTRODUCTION

Conversational agents or dialogue systems, also known as chatbots in the media and industry, have become very common in our daily lives. They can be found in our phones as personal assistants or on e-commerce websites as selling bots.

These systems are designed to communicate with humans in natural language text, speech, or both. Creating intelligent conversational agents remains an unresolved research problem that poses numerous challenges to the artificial intelligence community.

Through this work, we hope to identify the various existing algorithms for building chatbots. We also categorize and compare the dominant approaches based on the final use case, noting each approach's strengths and weaknesses. We hope to uncover issues related to this task that will assist researchers in deciding on future directions in conversational NLP. Although the first chatbots were created many years ago, attention has never been focused more on this area as it has in recent years. This can be explained by recent advancements in AI and NLP technologies, as well as the availability of data.

The current scientific and technological landscape is becoming crowded with the various methods for developing chatbots as in [1], [2], [3] and [4]. While there haven't been many resources up till now to help researchers and business focus on enhancing chatbot performance, there have been many recent successes in scaling reinforcement learning (RL) to complex sequential decision-making problems (e.g.: video games, self-driving cars, Chatbots.) which was sparked by the Deep Q-Networks algorithm [5] a combination of Q-learning, convolutional neural networks, and experience replay enabled it to learn how to play many Atari games at human-level performance from raw pixels. Many extensions have been proposed since then to improve its speed or stability.

In the next sections of this paper, we first studied the problematic and came up with ways to address these issues. Following that, we explored in-depth the state-of-the-art architecture of our proposed reinforcement learning model. We then proceeded to the experimental results where we compared other models to ours and their effectiveness of course.

## 2. RELATED WORK

Reinforcement Learning (RL) techniques are also based on value functions or policy search [6] which is typical of reinforcement learning (RL) techniques. Value functions have been used mostly in task-oriented conversation systems as in [7] and [8], whereas policy search has been used primarily in open-ended dialogue systems [9] [10], such chatbots. Given that chatbot systems employ unlimited action sets, task-oriented conversation systems use finite action sets, this is not surprising. As of now, policy search approaches for chatbots are favored, however it is unclear if this preference should continue given that these methods have issues with local rather than global optima, inefficiency, and large variation. As a result, this study examines the viability of a novel combination of reward value function for chatbots that has not, or at least not from the stand point of automatically generating reward sets, been investigated before.

In addition to deep RL, Seq2Seq models for conversation generations are also closely related techniques as in [11] and [12]. Because these approaches are often trained on millions of phrases, which implies enormous computing needs, they tend to be data-hungry. Although they may both be used to the same issue, in this research we concentrate on enhancing a Seq2Seq model using an RL-model employing a novel combination of reward functions, leaving the potential of other hyper-parameters for future work.

However, these related works concur that evaluation is a challenging task and that better evaluation measures are required [13]. This is further supported by Chia-Wei Liuet al.'s[14] research, which demonstrated that measures like Bleu and Meteor, among others, do not correlate with human judgements, lends more credence to this.

Regarding performance measures, depending on the application, deep RL dialogue agents may manually specify or learn their reward functions from dialogue data. Jiwei Li et al. [9] employ an adversarial strategy, where the discriminator is educated to differentiate between human and nonhuman sentences and utilize those scores to inform the generator's training. As in [15], they use evaluations from individuals' ratings to train a reward function. These linked studies are all neural-based, and there is no one reward function that would be the best to apply in upcoming (chitchat) chatbots.

This inspired us to propose a novel combination of three reward functions based on the flow of informality, coherence, and simplicity of response, which are three principles to sustain any human conversation. As well as employing more suitable performance metrics, such as the ones that are described below, which are practical since they simply need data in raw text and are simple to implement.

## 3. PROBLEMATIC

In general, a Sequence to sequence (Seq2Seq) model is composed of two recurrent neural networks (RNN) that serve as the encoder and decoder, respectively. Despite the fact that Seq2Seq models can generate grammatical and fluent responses as shown underneath.

A: Where are you going? (1)            A: how old are you? (1)
B: I'm going to the restroom. (2)      B: I'm 16. (2)

Figure 1.  Expected Seq2Seq conversation

The LSTM sequence-to-sequence (SEQ2SEQ) model is a type of neural generation model that maximizes the likelihood of producing a response given the previous dialogue turn. But they have been reported of having a diversity typically low because many responses are trivial or non-committal, such as" I don't know"," I'm sorry" or" I'm OK". This is known as the low-diversity problem. In Brief, a non-optimized Seq2Seq model: • Tend to generate highly generic responses. Can become trapped in an endless cycle of automatic replies.

**Baseline mutual information model (Li et al. 2015)**
A: Where are you going? (1)
B: I'm going to the restroom. (2)
A: See you later. (3)
B: See you later. (4)
A: See you later. (5)
B: See you later. (6)
...
...
A: how old are you? (1)
B: I'm 16. (2)
A: 16? (3)
B: I don't know what you are talking about. (4)
A: You don't know what you are saying. (5)
B: I don't know what you are talking about . (6)
A: You don't know what you are saying. (7)
...

Figure 2.  Non-optimized Seq2Seq model conversation Sample

## 4. PROPOSED MODEL

This section describes in depth the components of the proposed RL model. Two agents comprise the learning system. We use p to represent the first agent's sentences and q to represent the second agent's phrases. The two agents communicate with one another. A conversation is an alternate sequence of utterances produced by two agents: p1, q1, p2, q2, ..., pi, qi.

The produced sentences are considered as actions carried out by a policy set by a language model with an encoder-decoder recurrent neural network. As explained in section 4, the network parameters are optimized for Policy search, which is utilized to maximize the projected future reward. As a result, policy gradient approaches are more applicable for our circumstance.
In contrast to Q-learning [16], we can train the encoder-decoder RNN using MLE settings that already provide convincing answers; Before altering the target and adjusting toward a long-term reward-maximizing policy.

Q-learning on the other hand, explicitly predicts the future anticipated reward of each action, which might differ from MLE parameters and surpass the MLE objective by orders of magnitude, making them inappropriate for the initialization of our components (states, actions, rewards, and so on). The sequential decision problem is summarized in the subsections that follow.

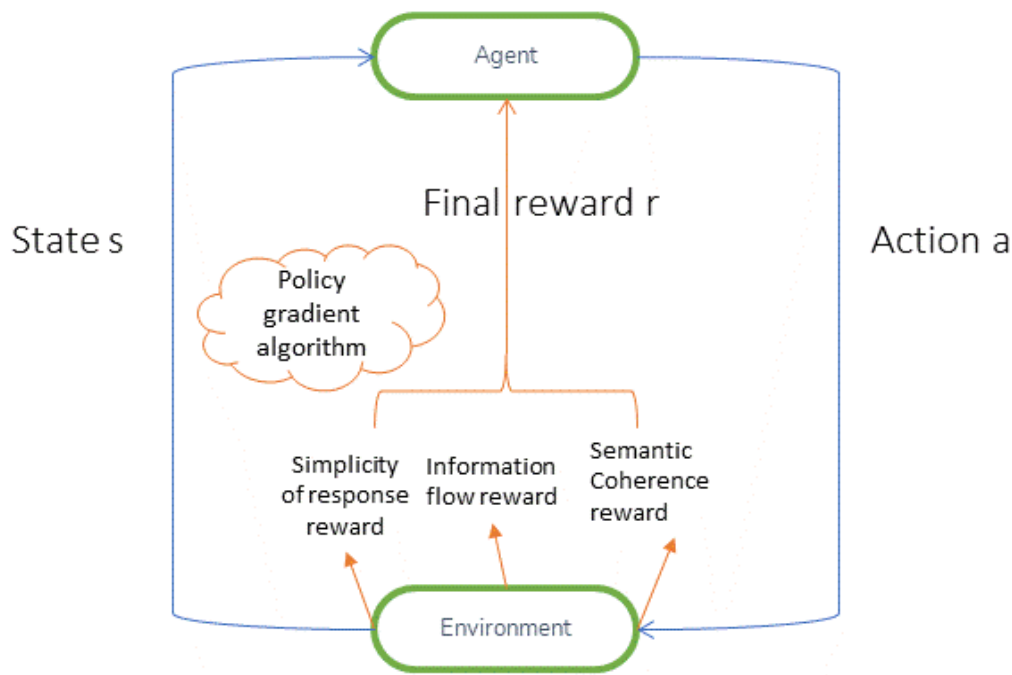

Figure 3. Proposed RL model schema.

### 4.1. Action

The dialogue utterance to generate is an action a. Because arbitrary-length sequences can be generated, the action space is indeed limitless.

## 4.2. State

The previous two dialogue turns $[p_i, q_i]$ denote a state. The dialogue history is then vectorized by feeding the concatenation of pi and qi into an LSTM encoder model, as described by Li et al. [17].

## 4.3. Reward

The reward obtained for each action is denoted by r. The 3 major factors that contribute to the success of a dialogue in this subsection, as well as how approximations to these factors can be operationalized in computable reward functions which are: Simplicity of response, Information Flow and Semantic Coherence.

### 4.3.1. Simplicity of Response

Simplicity of response is the avoidance of utterances with dull responses. There is a list of dull responses such as" I'm not sure what you're talking about,"," I have no idea," and so on which occurs very frequently in SEQ2SEQ models of conversations. We propose measuring response simplicity by calculating the negative log likelihood of responding to that utterance with a dull response. The following is the reward function:

$$r_1 = -\frac{1}{N_S}\sum_{s\in S}\frac{1}{N_s}\log p_{seq2seq}(s|a)$$

Where $N_s$ stands for the number of tokens in the dull response s and $N_S$ for the cardinality.
And $p_{seq2seq}(s|a)$ represents the likelihood output by SEQ2SEQ models.
Note that higher likelihood of a dull response results in lower reward.

### 4.3.2. Information Flow

Information flow is the resemblance in semantics between successive turns by the same agent. In order to maintain the pace of the dialogue and avoid any repetition in the sequences, we desire that each agent add new additional knowledge to the conversation at each turn. As a result, we want the semantic similarity between the same agent's subsequent turns should be punished. The reward is given by the negative log of the cosine similarity between $h_{pi}$ and $h_{pi+1}$:

$$r_2 = -\log\cos(h_{pi}, h_{pi+1}) = -\log\cos\frac{h_{pi}\cdot h_{pi+1}}{|h_{pi}||h_{pi+1}|}$$

Where $h_{pi}$ and $h_{pi+1}$ denote encoder representations for two consecutive turns $p_i$ and $p_{i+1}$.
Note that higher cosine similarity results in lower reward.

### 4.3.3. Semantic Coherence (Mutual Information)

Semantic Coherence is the avoidance of situations in which the generated responses are highly rewarded but are ungrammatical or incoherent. In order to evaluate the level of semantic coherence between generated questions and answers, we used a similarity evaluation method inspired by sentence-level similarity evaluation [18]. The following is the semantic coherence function:

$$r_3 = \frac{1}{N_a} \log p_{seq2seq}(a|q_i, p_i) + \frac{1}{N_{q_i}} \log p_{seq2seq}^{backward}(q_i|a)$$

Where $p_{seq2seq}(a|q_i, p_i)$ denotes the probability of generating action a, given state $[p_i, q_i]$.
And $p_{seq2seq}^{backward}(q_i|a)$ denotes the backward probability of generating the previous dialogue $q_i$ given action a.

Note that Lower Semantic Coherence score results in lower reward.

### 4.3.4. Final Reward Function

The total of all previous rewards is used to calculate the final reward for an action a.

$$r(a_i, [p_i, q_i]) = \lambda_1 r_1 + \lambda_2 r_2 + \lambda_3 r_3$$

where $\lambda_1 + \lambda_2 + \lambda_3 = 1$.
Since we wanted our model to focus on getting answers which are coherent and grammatically correct, due to the fact that without these two, generally the conversation with a real human would end soon if not instantly. As a result, we set $\lambda_3 = 0.5$ which means 50% of the final reward, giving semantic coherence much more importance than the other two (Information flow and Simplicity of response) which are only 25% each. As a result, we determined that $\lambda_1 = 0.25$, $\lambda_2 = 0.25$, and $\lambda_3 = 0.5$.

Note that after the agent completes each sentence, a reward is observed.

## 4.4. Policy

A policy is defined by its parameters and takes the form of an LSTM encoder-decoder. it's worth noting that a deterministic policy will always return the same action at a given state, in other words, it produces a discontinuous objective which would be difficult to optimize using gradient-based methods.

However, a stochastic policy (a probability distribution over actions given states) will generate a distribution probability on the actions. In our case, we employ policy gradient methods to identify parameters that result in a higher expected reward. The expected future reward is the goal to maximize:

$$J_{RL}(\theta) = E_{PRL(a_{1:T})}\left[\sum_{i=1}^{i=T} R(a_i, [p_i, q_i])\right]$$

where $R(a_i, [p_i, q_i])$ denotes the reward resulting from action ai.

## 4.5. Curriculum Learning Strategy

We use a curriculum learning strategy in which we start by simulating the dialogue for two turns and gradually increase the number of simulated turns. The candidate list's size increases exponentially as the number of candidates to be examined does. we generate no more than 5 turns. At each step of the simulation, five candidate responses are generated.

In brief, our RL model will look like something like this:
- State: $[p_i, q_i]$.
- Action: Generate a sequence. The action space is limitless.
- Reward: Simplicity of response, Information Flow, Semantic Coherence.
- The SEQ2SEQ model is similar to a policy gradients model based on the existing state of the world, it generates actions.
- SEQ2SEQ uses MLE to optimize parameters.
- RL optimizes the parameters based on the rewards.

## 5. SIMULATIONS

The main idea behind our approach is to simulate the process of two virtual agents talking to each other while exploring the state-action space so that our model can learn a policy that leads to the best-expected reward. In our paper today, we will compare 3 approaches:
- Supervised learning for Seq2Seq model.
- Semantic Coherence (Mutual Information) for pretraining policy model.
- Dialogue Simulation between Two Agents (based on our RL model).

### 5.1. Supervised Learning for Seq2Seq Model

For the first stage of training, we use the supervised SEQ2SEQ model to predict a generated target sequence given dialogue history [11]. The results of supervised model will be used for initialization later on. On the Open Subtitles dataset, which contains approximately 80 million source-target pairs, we trained a SEQ2SEQ model with attention [19]. The concatenation of two prior sentences was considered as a source input, while each turn in the dataset was treated as a target.

### 5.2. Semantic Coherence for Pretraining Policy Model

Samples from SEQ2SEQ models are frequently dull and generic, such as" I don't know" [17].As a result, we do not want to use pre-trained SEQ2SEQ models to initialize the policy model because this will result in a lack of diversity in the RL models' experiences.Li et al. [17] demonstrated that modelling semantic coherence between sources and targets significantly reduces the likelihood of producing dull responses and improves overall response quality. We now demonstrate how to obtain an encoder-decoder model that produces the most semantic coherence responses.

### 5.3. Simulation of a Conversation Involving Two Agents (Based on our RL Model)

We simulate conversations between the two virtual agents and have them talk to each other in turn. The simulation proceeds as follows: the first agent is fed a message from the training set in the first step. The agent vectorizes the input 6 message before beginning to decode it in order to generate a response output. Combining the current output of the first agent with the conversation history, the second agent changes the state by encoding the conversation history into a representation and generating responses with the decoder RNN, which are then sent again to the first agent, and the process is repeated again.

Putting everything together:
1. Supervised learning, training SEQ2SEQ model: (a) Not good for initializing policy.
2. Semantic Coherence for the pre-training policy model:
(a) Initialize with the model from (1).

(b) For each data point, there's an MLE loss, and a Semantic Coherence Score (reward).

(c) Consider a curriculum learning technique that excels at both.

- With the help of the simulated turns and rewards in order to maximize the expected future reward.

## 6. EXPERIMENTAL RESULTS

This section describes experimental results as well as qualitative analysis. Two automatic measures are used to evaluate the dialogue generation chatbot.: conversation's length (number of turns in the entire session) and diversity.

### 6.1. Dataset

The agent requires high-quality initial inputs for the dialogue simulation. An initial input of" why?" for example, is undesirable because it is unclear how the dialogue could proceed. To ensure that initial inputs are easy to answer, we will use a subset of 10 million conversations from the Open Subtitles dataset and extract 0.8 million sequences having the least chance of getting the reply" I have no idea what you're talking about.

### 6.2. Model Evaluation

It is difficult to evaluate dialogue systems. Metrics such as BLEU [20] and perplexity have been broadly used for evaluating dialogue quality as in [17], [11] and [21], but how well these automatic metrics are correlated with true response quality is widely debated as in [14] and [22]. We do not use BLEU or perplexity for evaluation because the proposed system's objective is the dialogue's long-term success, not anticipating the response with the highest likelihood. and its diversity. As a consequence, our RL model has boosted the baseline SEQ2SEQ model drastically as it is shown in Table 1.

Table 1. sampled answers produced by the proposed RL model and the semantic coherence models.

| Message Input | Semantic Coherence Model | Proposed RL Model |
|---|---|---|
| How old are you? | I'm 16. | I'm 16, why are you asking? |
| What's your full name? | I have no idea | What's yours? |
| I don't want to go home tonight. | Really? | Why? |
| Do you have any feelings for me? | I don't know what you are talking about. | I don't want to hurt your feelings. |
| How much time do you have here? | Not long enough. Sorry, sir. | Ten seconds. |
| Shall we get started? | Of course! | Yes. We've got a lot of work to do here. |
| Do you play football? | No, I don't | Yes. I love football! |
| We'd have to talk to him | He's a nice guy, after all. | About what? |
| How come you never say it? | Because I don't want to hurt you. | I don't think it's a good idea to say it |

### 6.2.1. Conversation's Length

The length of the simulated dialogue is the first metric we propose. When one of the agents starts producing dull responses like" I don't know," we say the dialogue has come to an end.

Three or two consecutive utterances from the same user have a high degree of overlap. The test set includes 1,000 input messages. We limit the number of simulated turns to less than 8 to reduce the risk of circular dialogues. Table 2 displays the results.

Table 2.  The average number of simulated turns from the Semantic Coherence model, the proposed RL model, and the standard SEQ2SEQ models.

| Models | # Of simulated turns |
|---|---|
| Sequence-to-sequence model | 2.76 |
| Semantic Coherence model | 3.54 |
| Proposed RL model | 4.63 |

### 6.2.2. Diversity

The number of distinct unigrams3 and bigrams4 in generated responses is used to calculate the degree of diversity. The value is scaled by the total amount of created tokens to prevent favoring extended sentences. as described by [17]. As a result, the resulting metric is a type-token ratio for unigrams and bigrams. The beam search we employed had a size of 10 to generate an answer to a given input message in both the standard SEQ2SEQ model and the proposed RL model. Table 3 displays the results.

Table 3.  Variance scores (type-token ratios) for the standard SEQ2SEQ model, Semantic Coherence model, and the provided RL model.

| Models | Unigram | Bigram |
|---|---|---|
| Sequence-to-sequence model | 0.0075 | 0.012 |
| Semantic Coherence model | 0.014 | 0.029 |
| Proposed RL model | 0.021 | 0.044 |

We discover that the proposed RL model produces more diverse outputs when compared to other models.

## 7. CONCLUSION AND FUTURE WORK

Given these findings, we believe that developing a natural conversational agent cannot be solved with a single simple model. As a result, we developed a reinforcement learning algorithm for the NLP chat generation bot by mimicking conversations between two agents, combining the strengths of neural SEQ2SEQ systems, and tuning its behavior using reinforcement learning.
Our model was able to generate utterances that maximize future rewards, effectively capturing the global properties of a good conversation. Although our model captures these global properties using very simple, operable heuristics, the model generates more diverse, interactive responses. Since practically every part of our system is a trainable machine learning model, more interactions and data are likely to result in significant system improvements.

Therefore, an exciting direction for future work is to extend offline RL model with more hyper-parameters and adopt other or even new performance metrics to address a broader range of human-computer interaction tasks, especially those with long-term dependencies as well as delayed rewards, where complex task goals can result in the creation of conversation that is eventually more beneficial to users.

# AUTHORS

**Ismail Jabri** is currently a second-year master's degree student specializing in Intelligent and Mobile systems at the Sidi Mohamed Ben Abdellah University. He received his bachelor's degree in business intelligence from Mohamed First University. His interests include natural language processing (NLP), machine and deep learning, the Internet of Things (IoT) and their real-world applications. He excited to continue exploring these areas and how they can be used to solve practical problems in a variety of industries.

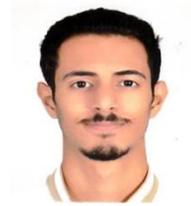

**Ahmed Aboulbichr** is a second-year master's degree student specializing in intelligent and mobile systems at Sidi Mohamed Ben Abdellah University. He received his bachelor's degree in web development from Higher Normal School of Tetouan, and have a strong interest in machine learning and deep learning. In his current studies, he has been focusing on exploring the applications of these technologies in various fields, in hope to contribute to the advancement of intelligent systems through his research and work.

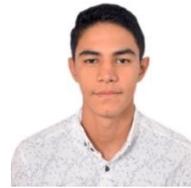

**Aziza El ouaazizi** received her PhD at Sidi Mohamed Ben Abdellah University in 2000. After working as a professor in Technical High School of Fes (2001), she is currently working for Accredited Professor in the Informatics at Sidi Mohamed Ben Abdallah University, Fez. She is also a permanent member of Artificial Intelligence Data Sciences and Emergent Systems Laboratory and an associate member of Engineering Science Laboratory. Her research interests include Machine and Deep Learning, Artificial Vision and image processing, Pattern Recognition, Data Analysis, Evolutionary Algorithms and their applications.

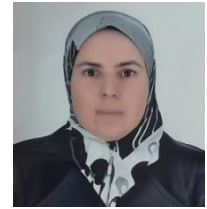